\documentclass[conference]{IEEEtran}
\IEEEoverridecommandlockouts
\usepackage{amsmath,amssymb,amsfonts}
\usepackage{algorithmic}
\usepackage{graphicx}
\usepackage{textcomp}
\usepackage{xcolor}
\usepackage{lipsum}
\usepackage{flushend}
\usepackage{tabularx}
\usepackage{multirow}
\usepackage{booktabs}
\usepackage[breaklinks]{hyperref}
\usepackage[linesnumbered,ruled,lined,commentsnumbered]{algorithm2e}
\usepackage{color,soul}
\usepackage[T1]{fontenc}
\usepackage[utf8]{inputenc}
\usepackage{authblk}
\usepackage[square, comma, sort&compress,numbers]{natbib}

\def\BibTeX{{\rm B\kern-.05em{\sc i\kern-.025em b}\kern-.08em
    T\kern-.1667em\lower.7ex\hbox{E}\kern-.125emX}}
\begin{document}

\title{BVI-Artefact: An Artefact Detection Benchmark Dataset for Streamed Videos\thanks{
\IEEEauthorrefmark{1}Equal contribution.

The research reported in this paper was supported by an Amazon Research Award, Fall 2022 CFP. Any opinions, findings, and conclusions or recommendations expressed in this material are those of the author(s) and do not reflect the views of Amazon. We also appreciate the funding from the China Scholarship Council, University of Bristol, and the UKRI MyWorld Strength in Places Programme (SIPF00006/1).}}

\author[1]{Chen Feng\IEEEauthorrefmark{1}}
\author[1]{Duolikun Danier\IEEEauthorrefmark{1}}
\author[1]{Fan Zhang}
\author[2]{Alex Mackin}
\author[2]{Andy Collins}
\author[1]{David Bull}
\affil[1]{\textit{Visual Information Laboratory, University of Bristol, Bristol, BS1 5DD, United Kingdom}}
\affil[1]{\textit {\{chen.feng, duolikun.danier, fan.zhang, dave.bull\}@bristol.ac.uk}}
\affil[2]{\textit{Amazon Prime Video, 1 Principal Place, Worship Street, London, EC2A 2FA, United Kingdom}}
\affil[2]{\textit {\{acmackin, accllin\}@amazon.co.uk}}

\maketitle
\begingroup\renewcommand\thefootnote{\IEEEauthorrefmark{1}}
\endgroup

\begin{abstract}
Professionally generated content (PGC) streamed online can contain visual artefacts that degrade the quality of user experience. These artefacts arise from different stages of the streaming pipeline, including acquisition, post-production, compression, and transmission. To better guide streaming experience enhancement, it is important to detect specific artefacts at the user end in the absence of a pristine reference. In this work, we address the lack of a comprehensive benchmark for artefact detection within streamed PGC, via the creation and validation of a large database, BVI-Artefact. Considering the ten most relevant artefact types encountered in video streaming, we collected and generated 480 video sequences, each containing various artefacts with associated binary artefact labels. Based on this new database, existing artefact detection methods are benchmarked, with results showing the challenging nature of this tasks and indicating the requirement of more reliable artefact detection methods. To facilitate further research in this area, we have made BVI-Artifact publicly available at \url{https://chenfeng-bristol.github.io/BVI-Artefact/} 
\end{abstract}


\section{Introduction}

Video streaming services are becoming more and more popular with increases in both the number of subscribers~\cite{Stoll_2023} and the diversity of streamed content. The streaming pipeline for professionally generated content (PGC) comprises multiple stages including acquisition, post-production, encoding, transmission and presentation \cite{bull2021intelligent};  each of these can introduce visual artefacts, resulting in reduced perceptual quality of the streamed videos. To deliver optimal user experience, it is important to identify and locate these artefacts in order to control the streaming quality.

Existing work on video quality assessment (VQA)~\cite{w:VMAF, danier2022flolpips, feng2022rankdvqa} has mainly focused on offering a single-dimensional prediction for the overall video quality -  either in a full-reference (FR) or no-reference manner (NR). Such an approach is limited in providing an underlying reason for the quality degradation. On the other hand, if we are able to recognise specific artefacts~\cite{goodall2018detecting}, this can be exploited in the streaming pipeline to monitor and fix these issues \cite{zhang2015perception}. 

Detecting artefacts in streamed videos in the absence of an artefact-free reference is a challenging task.  Among the limited work in this field, a significant contribution was reported in \cite{goodall2018detecting}, where the detection of eight common video artefacts in PGC streaming was studied. However, this work assumed the existence of a single type of artefact in each video, which is not realistic in most practical scenarios where artefacts generated at various stages of video streaming can co-exist and interact. This limitation has been observed in several other investigations~\cite{tandon2021cambi,zhao2023full} on artefact detection, which also consider only a single type of artefact in each video sequence. Other relevant work~\cite{wu2023towards} assesses various artefacts induced by acquisition and delivery processes. However, its main focus was user generated content (UGC), which exhibits characteristics that differ significantly from the PGC case (e.g., different distortions due to inadequate photography skills and unprofessional equipment).  We thus conclude that a realistic and comprehensive benchmark for artefact detection is needed for streamed PGC; this is the focus of this work.

% \begin{figure}[t]
%     \centering
%     \includegraphics[width=\linewidth]{figures/system.pdf}
%     \caption{The video streaming pipeline involving artefact detection. Detecting specific artefacts provides better guidance for technical systems on effectively improving streaming quality.}
%     \label{fig:system}
% \end{figure}

In this context, we propose the first public benchmark database for detecting artefacts in streamed PGC videos. To simulate real-world video streaming, we incorporated source videos containing five common inherent artefacts (\textit{motion blur, dark scene, graininess, aliasing and banding}), and employed various video processing methods to synthesise five further types of perceptual artefacts (\textit{blockiness, spatial blur, transmission errors, dropped frames and black frames}) induced in post-production, compression or transmission. As a result, our database is composed of 480 videos from 60 sources, each video containing up to six different artefacts accompanied with sequence-level binary labels (at frame level for some artefacts) for all artefacts. This database was then used to benchmark seven existing detection methods, and the results reveal the need for more accurate and robust artefact detection algorithms. 

% The rest of the paper is organised as follows. Section \ref{sec:database} describes the collection/creation of source sequences, the generation of each visual artefact and the annotation of artefact labels. Section \ref{sec:exp} presents the benchmarked detection methods and how their performance is evaluated. The detection results are then shown in Section \ref{sec:results} alongside the analysis and discussion. Finally, Section \ref{sec:conclusion} concludes the paper and outlines future works.

\section{Proposed Database}\label{sec:database}

\begin{figure*}[t]
    \centering
    \includegraphics[width=\linewidth]{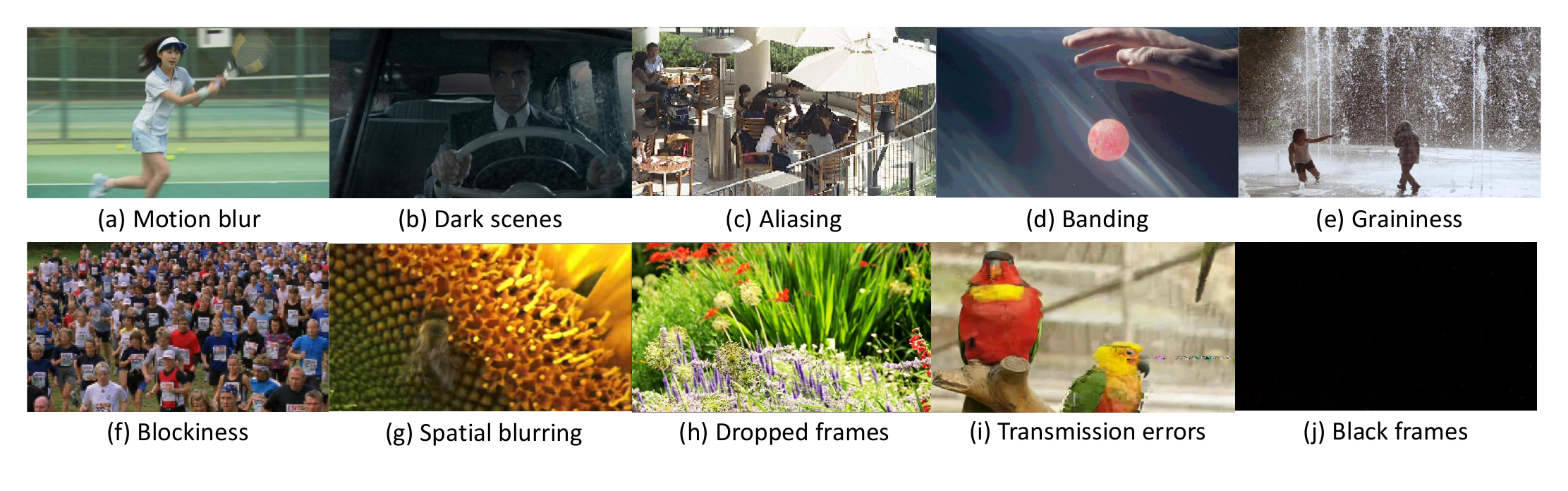}
        \vspace{-25pt}
    \caption{Examples of ten different visual artefacts included in the BVI-Artefact database.}
    \label{fig:sample}
\end{figure*}

\subsection{Source Sequence Collection}

Our BVI-Artefact database is derived from 60 PGC source sequences selected from five public video databases: BVI-HFR \cite{mackin2018study}, MCL-V \cite{mcl_V}, SVT Open Content Video Test Suite 2022 \cite{SVTTestSuite2022}, Netflix Open Content \cite{Netflixopen}, and Derf's collection from Xiph.org Video Test Media \cite{xiph}. These sequences contain natural scenes and objects, encompassing a large range of visual elements that include various landscapes, urban environments, activities and occurrences in the natural world, and typical content genres such as sports, drama, action, etc. They also exhibit diverse motions and texture types, including static textures, dynamic textures, translational motions, complex motions, various spatial structures and luminance uniform regions. The sequences are in YCbCr 4:2:0 format, with a range of resolutions from 4K (4096$\times$2160 and 3840$\times$2160) to HD (1920$\times$1080 and 1280$\times$720), frame rates (25, 50, 60, 120 fps) as well as bit depths (8 and 10 bits, non-HDR). Each sequence has a duration ranging from 5 to 10 seconds \cite{duration,moss2016support}. 

 In order to simulate PGC streaming in real-world applications, ten types of visual artefact common to PGC streamed content are introduced, as shown in  Fig.~\ref{fig:sample}. These artefacts are classified into two categories: \textbf{source} and \textbf{non-source} artefacts, where the former corresponds to artefacts arising from the acquisition process and post production, and the latter relates to artefacts typically generated during the delivery process including compression and transmission. 

\subsection{Source Artefacts}\label{sec:sourceartefacts}

\begin{figure}[t]
    \centering
    \includegraphics[width=\linewidth]{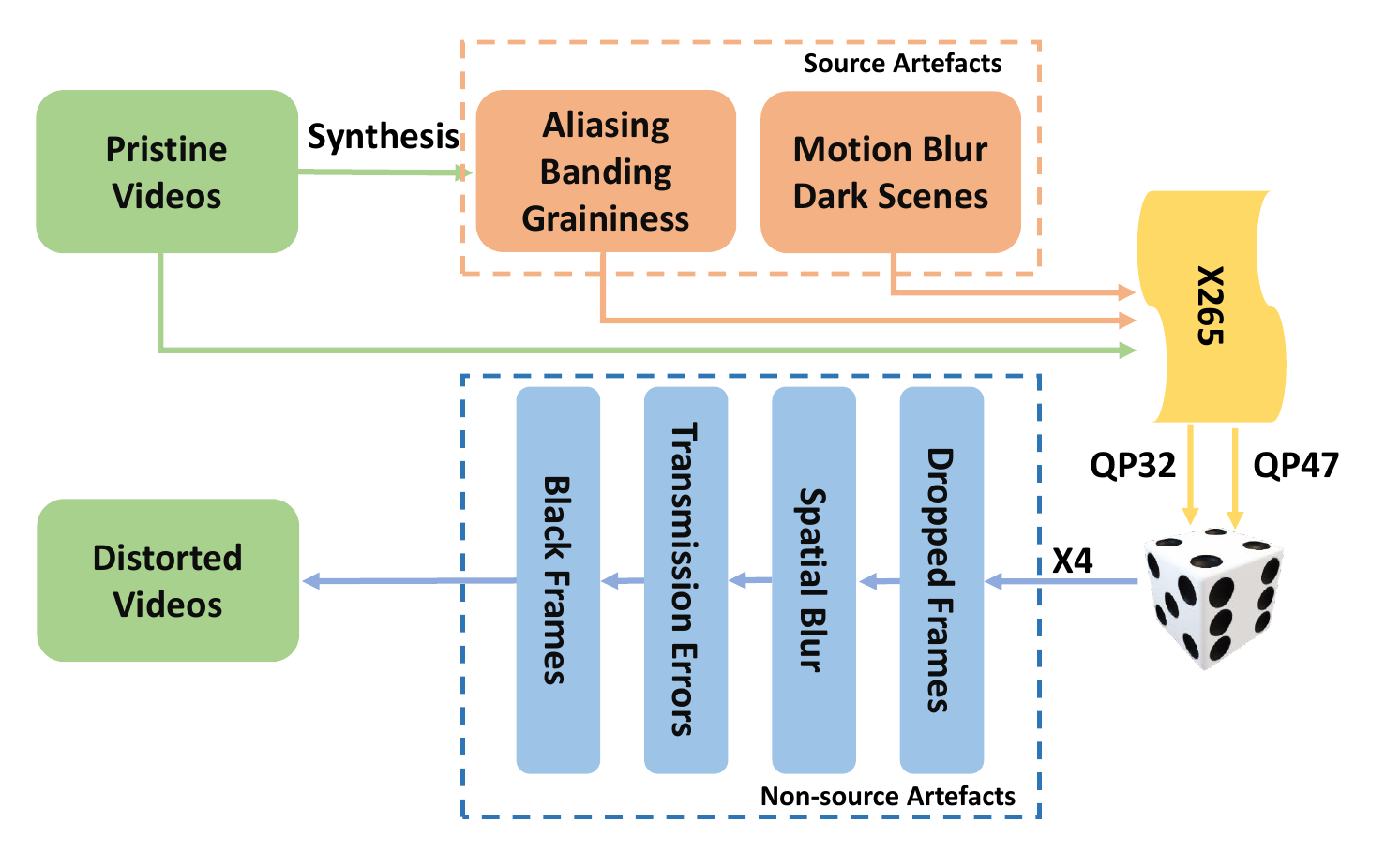}
        \vspace{-20pt}
    \caption{The workflow of the database generation.}
    \label{fig:datageneration}
        \vspace{-10pt}
\end{figure}

The process of introducing ten source or non-source artefacts is illustrated in Fig.~\ref{fig:datageneration}. First, we identify five common source artefacts usually induced during video acquisition and post-production: \textit{motion blur}, \textit{dark scenes}, \textit{graininess}, \textit{banding}, and \textit{aliasing}. While some of these artefacts are inherent to the original source videos, a few of them are synthesised based on the pristine source sequences, as described below.

\textbf{Motion blur} typically results from insufficient frame rates or relatively wide shutter angles, with areas corresponding to the rapidly moving parts of the scene with reduced high frequency energy. While motion blur can improve the perception of motion, it can also degrade visual quality when excessive~\cite{mackin2018study}. Considering the difficulty of synthesising realistic motion blur, ten source videos are collected with this artefact with various blur levels.

\textbf{Dark scenes}, when described as artefacts, reflect insufficient lighting (due to underexposure during acquisition) that leads to decreased perceptual quality. Although physics-based lighting degradation models~\cite{pisano1998contrast,ng2011total} exist for synthesising dark content, they generally yield reduced realism;  we therefore selected 10 natural dark sequences for this database with different darkness levels.

\textbf{Graininess} refers to the excessive visible noise in videos typically due to the nature of certain cameras (e.g., film camera) and/or high ISO values. Due to the limited availability of such content, to create grainy noise, we employed the technique in \cite{graininess} to add Gaussian noise (zero-mean, random standard deviations in a range of [0.1, 0.2]) to all frames of ten pristine source sequences.

\textbf{Banding} typically arises from the inadequate acquisition bit depth, which results in false contours in video frames (mainly manifesting as non-smooth colour changes). Banding can also be introduced and/or amplified due to the coarse quantisation in compression~\cite{krasula2022banding}. Here we followed the procedure in~\cite{goodall2018detecting} to stimulate banding artefacts by reducing the actual bit depth  of ten source sequences with a random bit-shift (3, 4, 5 or 6).

\textbf{Aliasing} artefacts often occur when video frames are spatially down-sampled without properly suppressing high-frequency signals beforehand. They typically appear as jagged edges and moir\'e patterns, leading to reduced visual quality. Aliasing is considered as a source artefact here, because these effects are typically introduced during content capture or at the initial processing stage before compression or transmission, particularly during spatial sampling. In this work, we synthesise aliasing artefacts in a random down-sampling range [2.0, 4.0] for  ten further pristine source videos through \cite{goodall2018detecting}.

In addition to these 50 source sequences, where every group of ten contains a unique source artefact, we also included ten pristine sources without any visual artefacts, forming a collection of 60 source sequences in total. 

\subsection{Non-source Artefacts}
\label{sec:nonsource}

As shown in Fig.~\ref{fig:datageneration}, each of the 60 source videos (pristine or with one source artefact) described above is compressed using an H.265~\cite{H265} codec (x.265~\cite{x265}, medium preset) at two different quantisation levels, resulting in 120 sequences, where half are heavily compressed (Quantisation Parameter, QP=47), and the other half are of acceptable (confirmed via manual inspection) quality (QP=32). It should be noted that we did not consider the case of uncompressed videos as these are not represented in real-world video streaming scenarios.

Five types of artefacts commonly introduced during down-stream video processing, compression and transmission are considered here: \textit{blockiness, spatial blur, transmission errors, dropped frames, and black frames}. As shown in Fig. \ref{fig:datageneration}, for each of the 120 compressed videos, we randomly synthesise \textit{dropped frames, spatial blur, transmission errors, black frames} sequentially, each with 50\% probability (i.e. each artefact has 50\% chance of being introduced). This random synthesis process has been repeated four times, resulting in four different versions for each of the 120 sequences, totalling 480 videos in the database. Below we describe the synthesis procedures for these non-source artefacts.

\textbf{Blockiness} typically results from commonly used video codecs that operate in a block-based manner. Since we have already encoded each video with x265~\cite{x265} (with QP values, 32 and 47),  we performed manual inspection to confirm that the QP=47 case does contain identifiable blockiness (see Section \ref{label}). 

\textbf{Spatial blur} refers to blurring artefacts generated by digital post-processing algorithms, e.g., smoothing filters (for deblocking or noise removal). However, these methods can also reduce the high frequency energy within a video frame, resulting in spatial blur~\cite{chakrabarti2010analyzing}. Here, following \cite{goodall2018detecting}, we synthesise spatial blur by convolving a Gaussian kernel (zero-mean, standard deviations in a random range of [0.1, 0.2]), and kernel size of 3$\times$3) with a randomly selected continuous segment (30\%) of frames in each video.

\textbf{Dropped frames} are typically due to data corruption or inconsistent network connection in video transmission, which often results in a severe impact on the user experience. To simulate dropped frames, we followed the method in~\cite{goodall2018detecting} to drop a number of randomly selected consecutive frames, and repeat the previous frame of the dropped frames until the total number of video frames remains unchanged.

\textbf{Black frames} are another common artefact in streamed content, which is usually caused by transmission errors, significantly interrupting the continuity of video playback. To simulate the occurrence of black frames, all the luma values is set to 16 (64 for 10-bit videos) and the chroma components are set to 128 (512 for 10-bit content) for 8-bit videos \cite{r:BT709}. The frame selection method is similar to that for synthesising dropped frames. It is noted here that black frames can be mixed with certain creative effects, e.g., scene fading. As we did not include sources with any scene cuts in this database, the investigation of this issue remains our future work.

\textbf{Transmission errors} are typically due to network congestion, interference, or protocol errors. These errors lead to data loss, significantly reduced perceptual quality, and interruptions in playback. To synthesise these conditions, following the approach in \cite{goodall2018detecting}, we use FFmpeg \cite{ffmpeg} to corrupt the bitstreams of H.265 videos, with a randomly adjusted corruption ratio. 

In summary, our database totals 480 videos, each with zero or one source artefact, and zero to five non-source artefacts.

\subsection{Artefact Label Annotation}
\label{label}

For each of the 480 sequences, we provide a binary label for each of the ten artefact types, indicating their presence. These labels were first annotated according to the artefact generation methods described above, and then confirmed through manual inspection to ensure that the labelled artefacts are indeed recognisable. As a result, for each source artefact, there are 80 (i.e. 16.7\%) positive videos (i.e. the artefact is visible)\footnote{This does not affect the use of this database for source artefact detection, as described in Section \ref{sec:benchmarking}}, and for each non-source artefact, there are around 240 (i.e. 50\%) positive samples. Additionally, for some non-source artefacts (i.e. \textit{spatial blur, dropped frames, black frames}), we also provided frame-level binary annotation.

\begin{figure*}[t]
    \centering
    \includegraphics[width=0.195\linewidth]{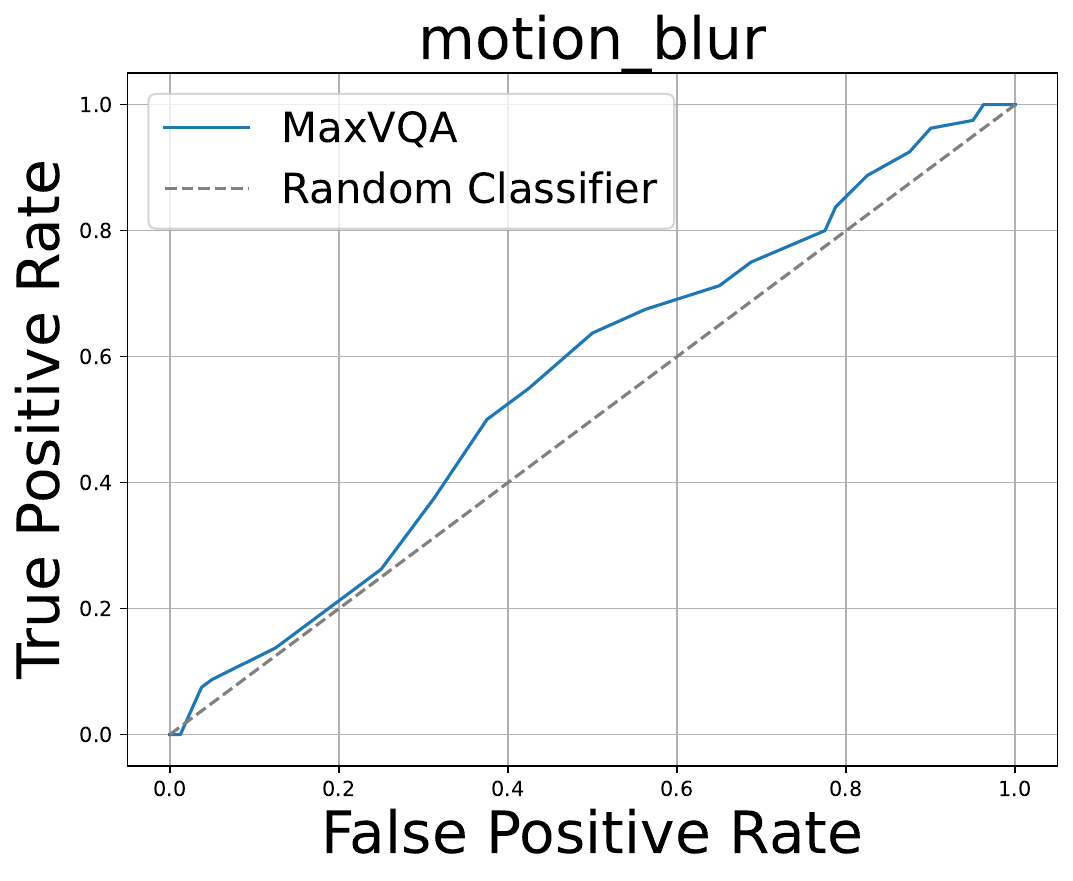}
    \includegraphics[width=0.195\linewidth]{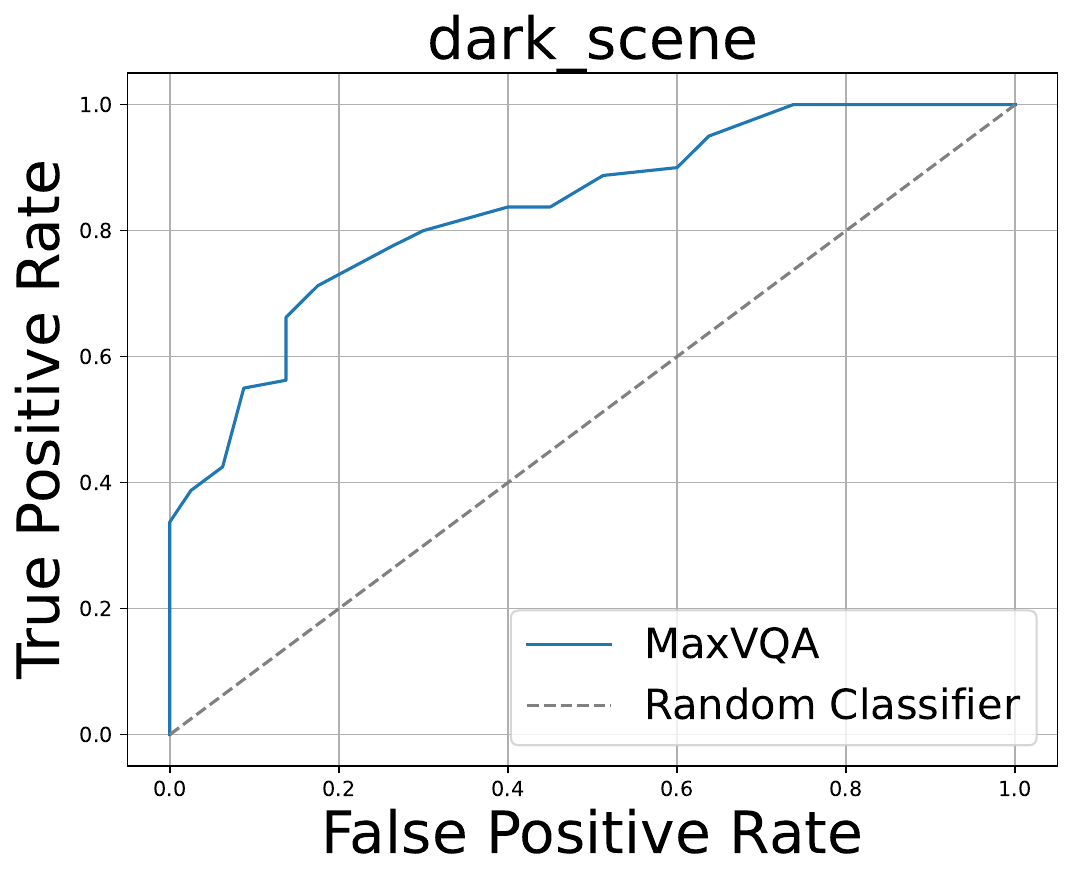}
    \includegraphics[width=0.195\linewidth]{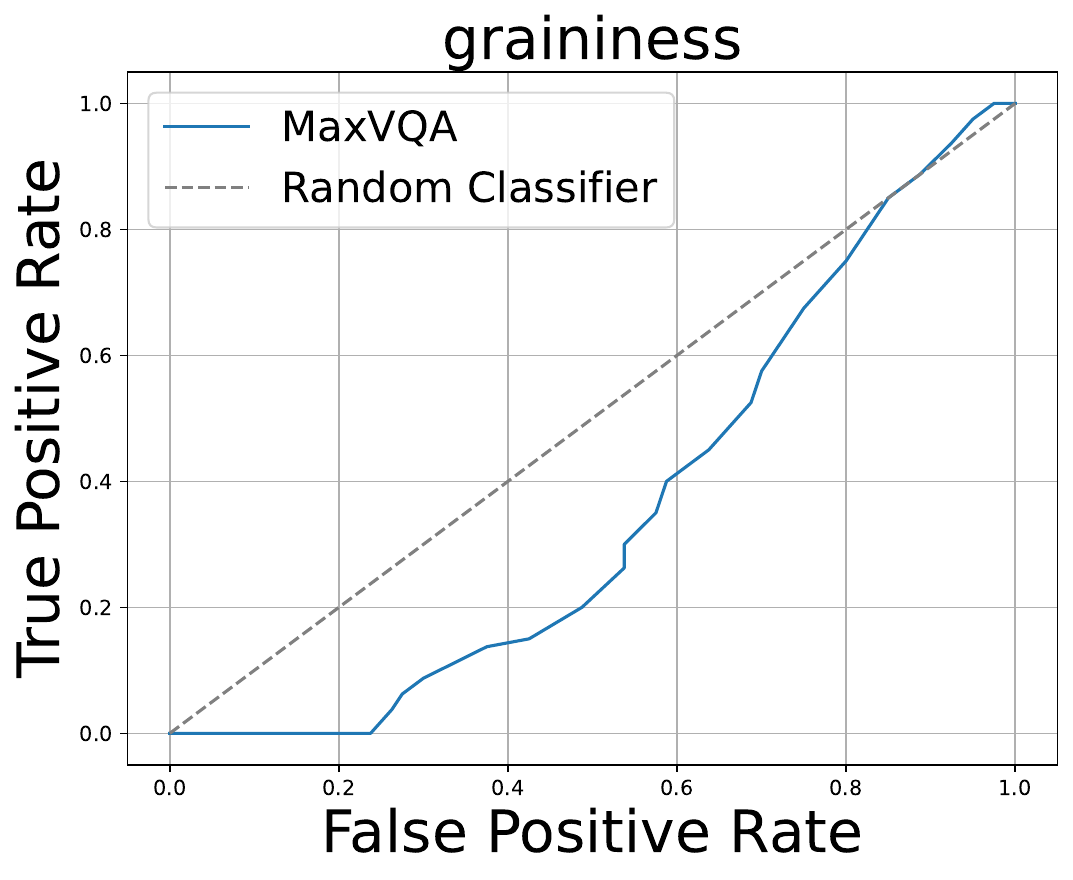}
    \includegraphics[width=0.195\linewidth]{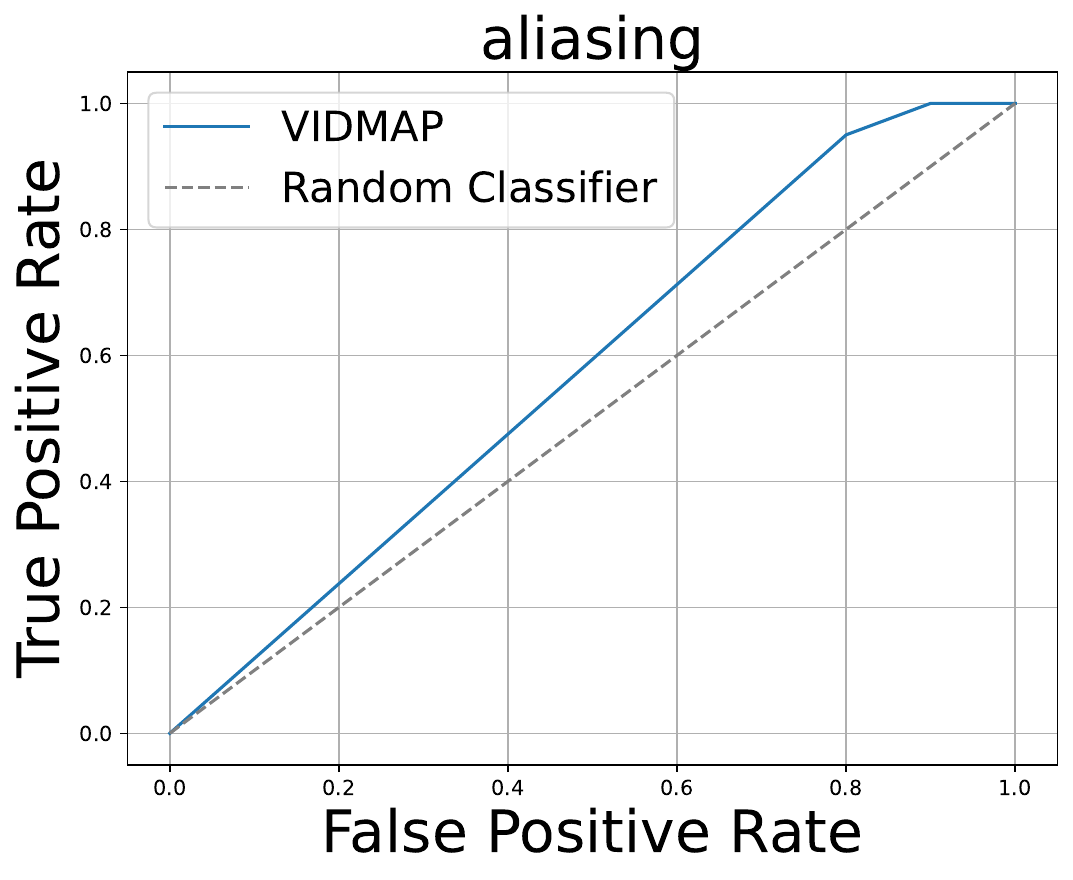}
    \includegraphics[width=0.195\linewidth]{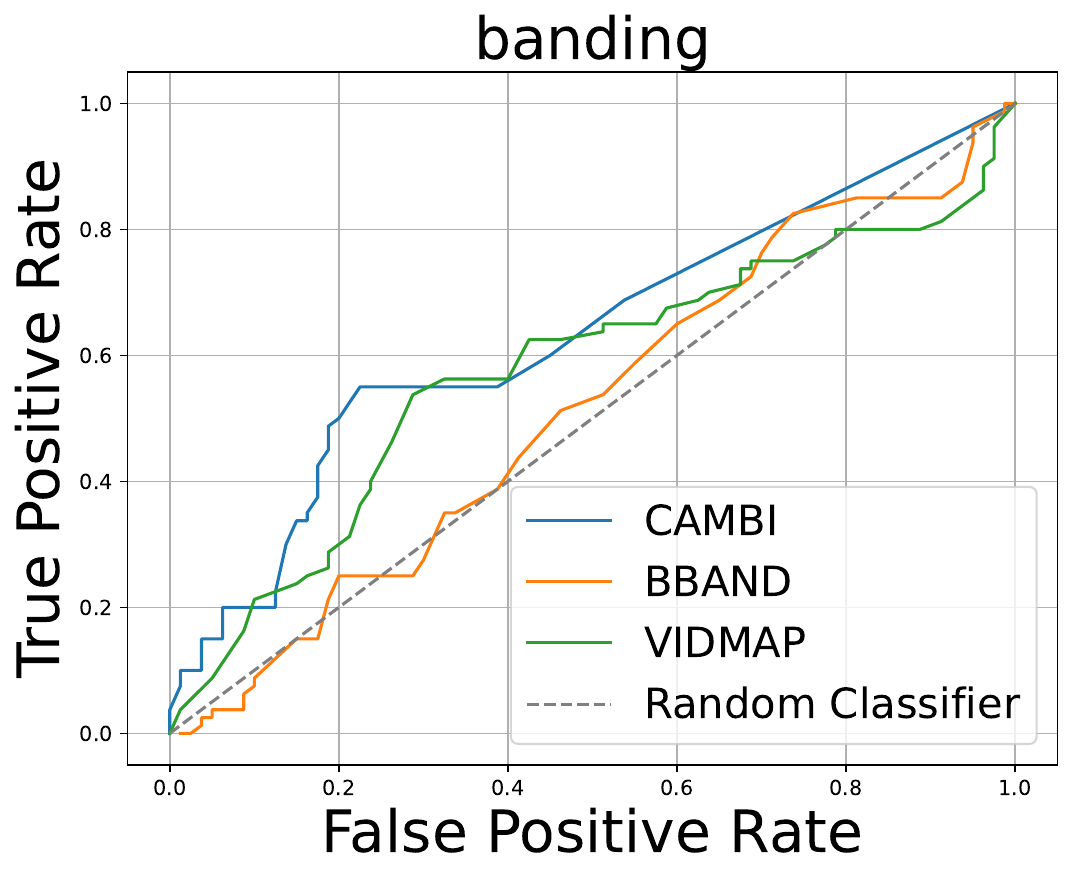}\\
    \includegraphics[width=0.195\linewidth]{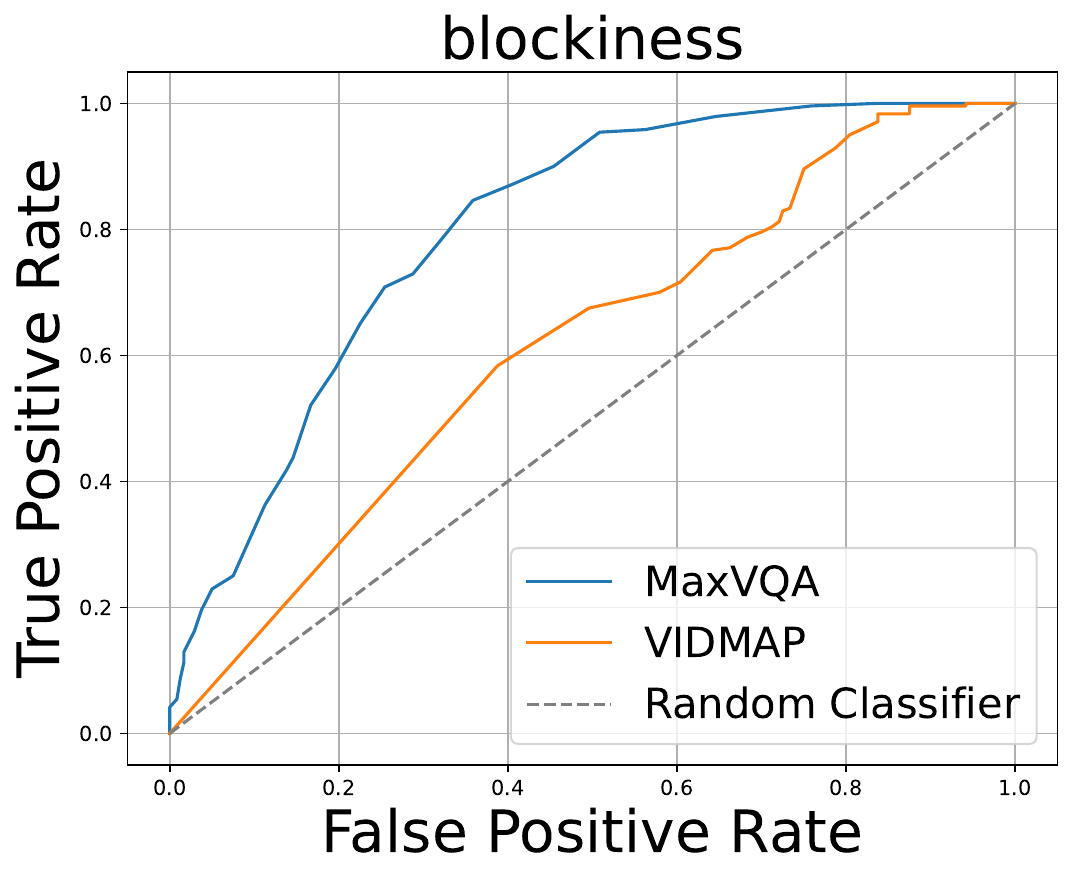}
    \includegraphics[width=0.195\linewidth]{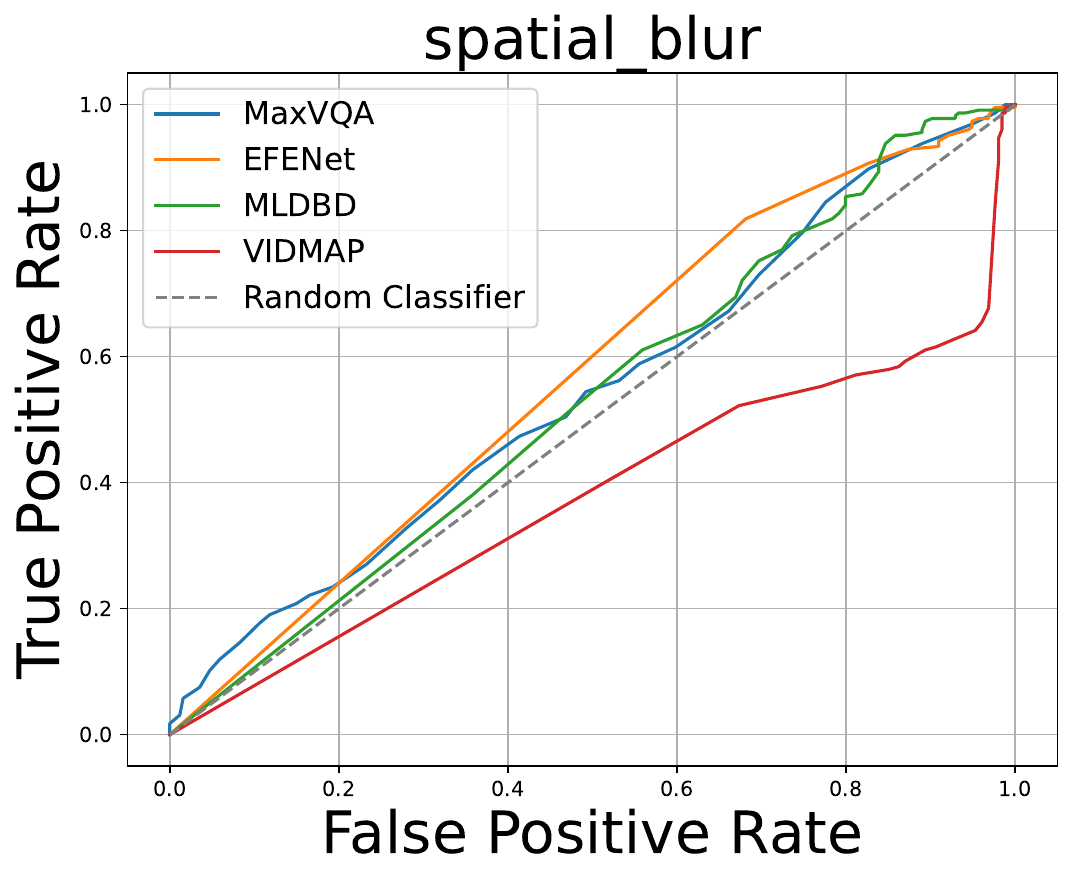}
    \includegraphics[width=0.195\linewidth]{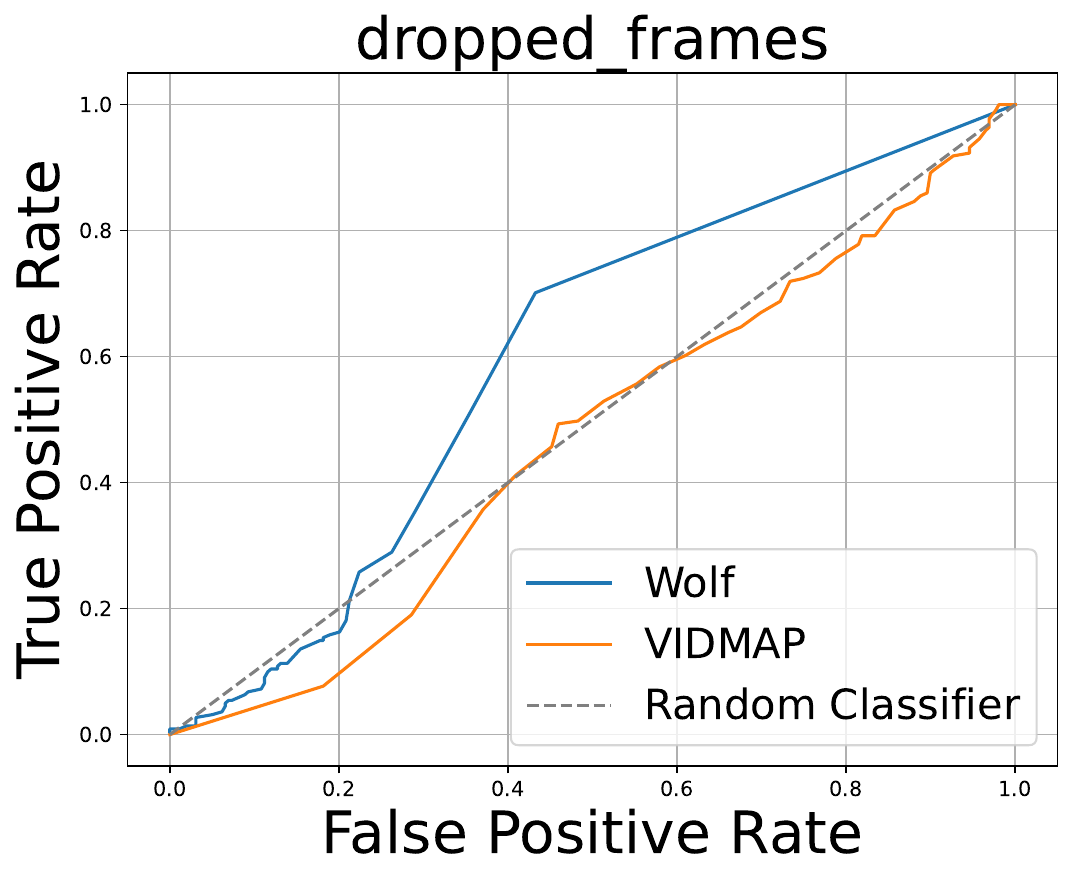}
    \includegraphics[width=0.195\linewidth]{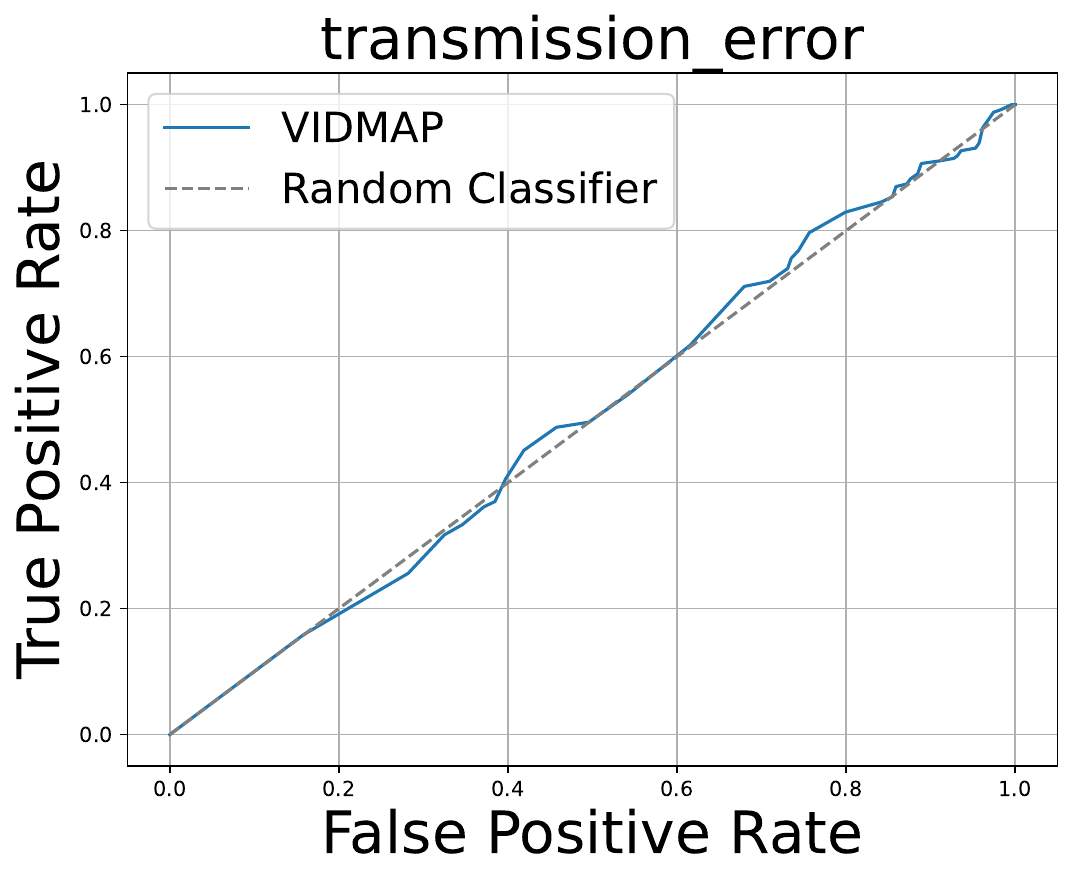}
    \caption{The ROC curves for different artefact detection.}
    \label{fig:roc}
    \vspace{-1em}
\end{figure*}

\begin{table}[t]
\centering
\caption{Benchmark results on various artefact detection tasks.}
\label{tab:results}
\begin{tabular}{r|l|c|c|c}
\toprule
\textbf{Artefact}                        & \textbf{Method}      & \textbf{Acc. (\%)} $\uparrow$ & \textbf{F1} $\uparrow$ & \textbf{AUC} $\uparrow$ \\ \midrule
Motion blur                     & MaxVQA      &     51.88&0.68&0.56 \\ \midrule
Dark scene                      & MaxVQA      &    73.13 &0.67&0.84 \\ \midrule   
Graniness                       & MaxVQA      &    38.75 &0.16&0.36 \\ \midrule
Aliasing                        & VIDMAP      &    50.00 &0.67&0.58 \\ \midrule
\multirow{3}{*}{Banding}        & VIDMAP      &    56.25 &0.59&0.58 \\
                                & CAMBI       &61.88     &0.53&0.63 \\
                                & BBAND       &50.00     &0.44&0.51 \\ \midrule
\multirow{2}{*}{Blockiness}     & MaxVQA      &64.58     &0.55&0.80 \\
                                & VIDMAP      &54.38     &0.69&0.61 \\ \midrule
\multirow{4}{*}{Spatial blur}   & MaxVQA      &53.54     &0.40&0.54 \\
                                & VIDMAP      &47.29     &0.64&0.38 \\
                                & EFENet      &47.08     &0.64&0.57 \\
                                & MLDBD       &49.58     &0.65&0.53 \\ \midrule
\multirow{2}{*}{Dropped frames} & VIDMAP      &45.42     &0.59&0.47 \\
                                & Wolf et al. &51.67     &0.18&0.60 \\ \midrule
Transmission error              & VIDMAP      &51.04     &0.65&0.50 \\ \bottomrule   
\end{tabular}
\vspace{-1em}
\end{table}

\subsection{Using BVI-Artefact for Benchmarking}
\label{sec:benchmarking}

Given the labels described above, the BVI-Artefact database can be used as a test set to evaluate the performance of artefact detection methods. Since we formulate the artefact detection problem as a binary classification task for each artefact, it is important to balance the positive and negative classes in the test set for each artefact. It is noted that our artefact generation process  naturally achieve such balance for each of the non-source artefacts, so the entire database is suitable for detection method benchmarking. For each source artefact, we take the 80 videos derived from the corresponding (to this artefact) source videos as positive samples, and randomly select another 80 videos from the rest of the database to form the negative class. As a result, we produce a subset of 160 videos for each source artefact to avoid class imbalance. These subsets are then fixed and used to evaluate the corresponding artefact detection methods. It is noted that the test data for each artefact may also contain various other artefacts which can interact with each other, and this makes BVI-Artefact a more practical and comprehensive benchmarking database. To facilitate further research, we have made the database (as well as the meta data for the subsets) publicly available.

\section{Experimental Configuration}
\label{sec:exp}

In this section, we describe the experimental configuration for benchmarking existing artefact detection methods on the proposed database.

\subsection{Benchmarked Methods} 

Seven existing artefact detection methods with publicly-available source code (and pre-trained model weights for deep learning models) have been evaluated on BVI-Artefact. These include two methods that can detect multiple artefacts, VIDMAP~\cite{goodall2018detecting} and MaxVQA~\cite{wu2023towards}, where the latter is designed for UGC videos. For banding artefacts, we evaluate CAMBI~\cite{tandon2021cambi} and BBAND~\cite{tu2020bband}, both inspired by the characteristics of the Human Vision System. For detecting spatial blur, we consider two state-of-the-art blur detection methods based on deep learning, EFENet~\cite{zhao2021defocus} and MLDBD~\cite{zhao2023full}. Regarding the detection of dropped frames, a classic method based on frame differencing proposed in~\cite{wolf2008no} has been evaluated. For \textit{black frames}, we did not find any specific detection methods, though for our database (without scene fading) simple thresholding could provide satisfactory results. However, the inclusion of this artefact is still important as it can interact with other artefacts, affecting their detection accuracy.

\subsection{Evaluation Metrics}
The task is formulated as binary classification, where given a (streamed) video, the goal here is to obtain a binary label for each artefact indicating its existence. Therefore, it is important to assess the accuracy (Acc.) of the detection. We follow the approach in~\cite{goodall2018detecting} to report the F1 score for all detection methods. Most benchmarked methods either output an overall probability or a score for a video, which is then converted to a binary prediction via thresholding. Here we use the original default threshold for each method to calculate the accuracy and F1 score values. We further evaluate these methods by varying such thresholds and drawing ROC (receiver operating characteristic) curves, from which the AUC (area under curve) values are obtained to indicate the overall performance. Regarding the detection of spatial blur, the evaluated methods (MLDBD and EFENet) predict a binary map for each frame to indicate the blurry pixels, instead of an overall index for a given video. To convert these maps to a video-level binary label, we take the average of all binary maps predicted for a video to obtain the percentage of pixels reckoned blurry by these methods, and consider the prediction positive (i.e. spatial blur does exist) if the average exceeds a default threshold of 30\% (See Section~\ref{sec:nonsource} \textbf{Spatial Blur}). For the calculation of AUC scores, this threshold is varied between 0-100\%.

\section{Results and Discussion}
\label{sec:results}

% 1. overall performance not good (best and worst)
% 2. spatial blur desiengd for images
% 3. challenging many not good because no consider interaction
TABLE \ref{tab:results} and Fig.~\ref{fig:roc} summarise the benchmark results for seven artefact detection methods on the proposed BVI-Artefact database. It is observed that firstly, the overall performance of the tested methods are not satisfactory, being similar to a random classifier in many cases. This can be mainly due to the fact that multiple artefacts often co-exist in the videos of our database, while such interaction between artefacts is not considered during the design of most of these detection methods, where often a single type of artefact is assumed. Additionally, the two \textit{spatial blur} detection methods, EFENet and MLDBD, are originally designed for single images so detecting spatial blur in videos can be challenging for them. Secondly, the UGC-specific method MaxVQA shows relatively good performance on detecting \textit{dark scene} and \textit{blockiness}, achieving AUC scores of 0.84 and 0.80 respectively. This can imply certain level of similarity in the manifestation of these artefacts in UGC and PGC content.

Overall, these results confirm that it remains a challenging task to detect the visibility of specific artefacts in the practical streaming scenario where various artefacts can co-exist and interact with each other. BVI-Artefact serves as a useful benchmarking platform to facilitate the development of better artefact detection methods.

\section{Conclusions} \label{sec:conclusion}

In this paper, we present BVI-Artefact, the first public benchmark database for artefact detection in streamed professionally generated content (PGC). This includes ten common visual artefacts and allows the inter-play of different artefacts to simulate practical streaming scenarios. We perform benchmarking for existing artefact detection methods on the proposed database and the results indicate the need for more robust and accurate detection methods for streamed PGC videos. Future work should study the influence of perceived artefacts on visual quality.

\vspace{5pt}
\small
\bibliographystyle{IEEEtran}
\bibliography{egbib}

\end{document}